\documentclass[nohyperref]{article}

\usepackage{microtype}
\usepackage{graphicx}
\usepackage{subfigure}
\usepackage{booktabs} 
\usepackage{hyperref}
\usepackage{xspace}
\usepackage{multirow}

\usepackage[accepted]{icml2022}

\usepackage{amsmath}
\usepackage{amssymb}
\usepackage{mathtools}
\usepackage{amsthm}

\usepackage[capitalize,noabbrev]{cleveref}

\theoremstyle{plain}

\theoremstyle{definition}

\theoremstyle{remark}

\newcommand{\proposedregmethod}{\textit{FedSwitch}\xspace}
\newcommand{\vanillaema}{TS-Server EMA\xspace}
\newcommand{\freqema}{TS-Client EMA\xspace}

\usepackage[textsize=tiny]{todonotes}

\newcommand{\subtitle}{Federated Semi-supervised Learning with Teacher-Student EMA}

\icmltitlerunning{\subtitle}

\begin{document}

\twocolumn[
\icmltitle{When does the student surpass the teacher? \\ \subtitle}



\icmlsetsymbol{equal}{*}

\begin{icmlauthorlist}
\icmlauthor{Jessica Zhao}{equal,meta}
\icmlauthor{Sayan Ghosh}{equal,meta}
\icmlauthor{Akash Bharadwaj}{meta}
\icmlauthor{Chih-Yao Ma}{meta}
\end{icmlauthorlist}

\icmlaffiliation{meta}{Meta Platforms, Inc., United States}

\icmlcorrespondingauthor{Jessica Zhao}{jessicazhao@meta.com}
\icmlcorrespondingauthor{Sayan Ghosh}{sayanghosh@meta.com}

\icmlkeywords{Semi-supervised Learning, Federated Learning, Pseudo-\\
 Labeling, teacher-student Models, Exponential Model Averaging}

\vskip 0.3in
]



\printAffiliationsAndNotice{\icmlEqualContribution} 

\newcommand {\jessica}[1]{{\color{orange}\textbf{Jessica: }#1}\normalfont}
\newcommand {\kevin}[1]{{\color{red}\textbf{Kevin: }#1}\normalfont}
\newcommand {\sayan}[1]{{\color{blue}\textbf{Sayan: }#1}\normalfont}
\newcommand {\akash}[1]{{\color{blue}\textbf{Akash: }#1}\normalfont}

\begin{abstract}
Semi-Supervised Learning (SSL) has received extensive attention in the domain of computer vision, leading to development of promising approaches such as FixMatch. In scenarios where training data is decentralized and resides on client devices, SSL must be integrated with privacy-aware training techniques such as Federated Learning. We consider the problem of federated image classification and study the performance and privacy challenges with existing federated SSL (FSSL) approaches.
Firstly, we note that even state-of-the-art FSSL algorithms can trivially compromise client privacy and other real-world constraints such as client statelessness and communication cost. Secondly, we observe that it is challenging to integrate EMA (Exponential Moving Average) updates into the federated setting, which comes at a trade-off between performance and communication cost.
We propose a novel approach \proposedregmethod, that improves privacy as well as generalization performance through Exponential Moving Average (EMA) updates. \proposedregmethod utilizes a federated semi-supervised teacher-student EMA framework with two features - \textit{local teacher adaptation} and \textit{adaptive switching between teacher and student for pseudo-label generation}. Our proposed approach outperforms the state-of-the-art on federated image classification, can be adapted to real-world constraints, and achieves good generalization performance with minimal communication cost overhead.
\end{abstract}




\newcommand\ifEMA[2]{\ifnum \value{EMA}>0{#1}\else {#2}\fi}
\newcommand{\modelparam}{\theta}
\newcommand{\student}{S}
\newcommand{\teacher}{T}

\newcounter{LaS}
\newcounter{EMA}
\newcounter{parallel}
\newcommand\ifLaS[2]{\ifnum \value{LaS}>0{#1} \else {#2}\fi}
\newcommand\ifparallel[2]{
    \ifLaS{\ifnum \value{parallel}>0{#1} \else {#2}\fi}{}
}

\newcommand{\suploss}{\mathcal{L}_{s}}
\newcommand{\unsuploss}{\mathcal{L}_{u}}
\newcommand{\kl}{D_\text{KL}}
\newcommand{\alphakl}{\beta}

\section{Introduction}
Semi-supervised Learning (SSL) has been successfully applied to many machine learning applications such as image classification, where there is potential for leveraging large amounts of unlabeled data for training.
Motivated by privacy concerns, Federated Learning (FL) was proposed by McMahan et al.~\cite{FedAvg} to train machine learning models on data distributed across several client devices without moving data off-device. 
The FL setting presents unique challenges to machine learning applications, such as client non-IIDness, where clients' data is sampled from different distributions~\cite{karimireddy2020scaffold}, scalability, and non-uniform selection of clients.


Federated Semi-Supervised Learning (FSSL) is an emerging area of research, which considers scenarios where unlabeled data is available client-side, such that one can apply existing SSL methods to leverage this data at the client devices to train and further improve the performance of the model.
Several FSSL approaches have been proposed, such as \textit{FedMatch}~\cite{FedMatch} and \textit{FedRGD}~\cite{FedGRD}, which leverage unlabeled data to train an image classifier in the federated setting. 
Despite achieving good generalization performance on experimental settings when assuming IIDness, low number of clients etc.,
some existing work fails to satisfy the same client privacy guarantees as traditional FL methods or assumes stateful clients participating in each round. 
For example, some of these approaches such as FedMatch~\cite{FedMatch} relax their design assumptions leading to trivial privacy compromises and sharing clients' models with each other. These assumptions are critical in industrial deployments, where clients are stateless and subject to tight privacy controls.
These are both important considerations when designing practical FSSL algorithms~\cite{kairouz2021advances}, and we provide a detailed discussion focusing on \textit{privacy leakage} and \textit{client state} in Section~\ref{real-world-design}.

Teacher-student EMA~\cite{tarvainen2017mean, liu2021unbiased} is a popular paradigm in semi-supervised learning, where there is a teacher model generating pseudo-labels for unlabeled data, while being updated slowly through a exponential weighting mechanism. The advantage of this approach is that the teacher model is relatively robust to changes in the student model and can improve pseudo-label quality, which translates to better generalization performance. While there has been prior work on federated teacher-student approaches~\cite{DBLP:journals/corr/abs-2012-03292}, to the best of our knowledge, there has been no analysis of the design considerations and performance tradeoffs of such approaches applied to FSSL. In a centralized server setting, EMA updates occur after every batch and while slowly updating the teacher, keeps it synchronized with the student model. In federated learning, an implementation of this EMA update scheme would be non-trivial. For example, some key design decisions are the placement of the teacher and student models on the server and individual clients and how frequently student model weights are communicated to the teachers for EMA updates in addition to the standard FL updates. In this paper, we discuss potential approaches along with trade-offs between communication cost and generalization performance.



We find that depending on the setting and stage of training, the pseudo-labels generated by the teacher may be of higher quality than the pseudo-labels generated by the teacher.
We propose \proposedregmethod---a novel variant on teacher-student EMA techniques for federated SSL, which dynamically switches between using the teacher and the student to generate pseudo-labels. 
Through extensive experimentation, we show that \proposedregmethod achieves state-of-the-art performance
on the CIFAR-10 and Fashion MNIST datasets in a federated setting. 
The improved training properties of \proposedregmethod{} are further manifested in a smoother convergence curve, as well as reduced model variance. 
We plan to publicly release our code to facilitate future research. To summarize, our key contributions are as follows:

\begin{enumerate}
    \item We study the limitations of current FSSL (Federated Semi-supervised Learning) techniques in practical environments and the complexities encountered when designing teacher-student EMA approaches in federated settings.
    \item
    We propose \proposedregmethod, a federated teacher-student EMA model with a novel mechanism, relying on local teacher adaptation and leveraging either the teacher or student model for pseudo-label generation based on a KL-Divergence metric. 
    \item We demonstrate that \proposedregmethod\ achieves state-of-the-art generalization performance on the CIFAR-10 dataset and improves on existing approaches such as FedMatch~\cite{FedMatch}, FedRGD~\cite{FedGRD} and FedProx-FixMatch~\cite{sohn2020fixmatch}. 
\end{enumerate}

\section{Related Work}
\label{related-works}
\subsection{Federated Learning}
Since the introduction of FedAvg~\cite{FedAvg}, which generalized SGD to the collaborative learning setting where data is distributed across multiple devices, several improvements~\cite{kairouz2021advances} have been proposed, such as FedProx~\cite{FedProx}, which generalizes and improves upon FedAvg by adding a proximal term $\mu$ to the local objective. 
Furthermore, several methods have been proposed to address specific issues in federated learning, such as improving privacy via differentially private training~\cite{dwork2014algorithmic,wei2020federated} or reducing communication costs~\cite{konevcny2016federated}.

\subsection{Semi-Supervised Learning}
The majority of the recent SSL methods can be roughly categorized by their two main components: (1) input augmentations and perturbations, and (2) consistency regularization.
The main motivation of these methods is to regularize the model to be invariant and robust to data or feature augmentations, which they achieve by enforcing the output predictions from the original and augmented inputs to be consistent with each other.
For example, in image classification, existing approaches apply various data augmentation strategies that generate different transformations of semantically identical images~\cite{berthelot2019mixmatch,laine2016temporal,sajjadi2016regularization,tarvainen2017mean}, perturb the input images along the adversarial direction~\cite{miyato2018virtual,yu2019tangent}, mix input images to generate augmented training data and annotations~\cite{zhang2018mixup,yun2019cutmix,guo2019mixup,hendrycks2020augmix}, or learn augmented prototypes in feature space instead of image space~\cite{kuo2020featmatch}.

\subsection{Federated Semi-Supervised Learning}
FedMatch~\cite{FedMatch} is one of the very first works in FSSL, in which the authors improve over a na\"ive combination of FL and SSL 
by introducing an inter-client consistency loss to regularize models learned at multiple clients to make similar predictions. However, this approach violates privacy constraints in practice. Another approach, FedRGD~\cite{FedGRD} proposes a grouping-based averaging of client updates in place of FedAvg-based averaging to reduce the gradient diversity of models and improve generalization performance.

\begin{table*}[t]
\centering
\tabcolsep=0.1cm
\caption{Properties of existing FSSL approaches in terms of providing privacy guarantees similar to standard supervised FL (as in FedAvg) and assumption of client statelessness in their design. Note that FedMatch does not adequately provide client privacy guarantees and FedSEAL/FedCon do not assume stateless clients. None of the existing FSSL approaches consider teacher-student EMA modeling to improve performance.
}
\label{tab:design-prop}
\begin{tabular}{lcccccc}
\hline
\multicolumn{1}{c}{} & FedMatch & FedRGD & FedSEAL & FedCon & \proposedregmethod \\ 
\multicolumn{1}{c}{} & \cite{FedMatch} & \cite{FedGRD}  & \cite{bian2021fedseal} & \cite{long2021fedcon} & (ours)\\ 
\hline
Privacy & \multicolumn{1}{c}{$\times$} & \multicolumn{1}{c}{\checkmark}  & \multicolumn{1}{c}{\checkmark} & \multicolumn{1}{c}{\checkmark} & \multicolumn{1}{c}{\checkmark} \\ \hline
Statelessness & \multicolumn{1}{c}{\checkmark} & \multicolumn{1}{c}{\checkmark}  & \multicolumn{1}{c}{$\times$} & \multicolumn{1}{c}{$\times$} & \multicolumn{1}{c}{\checkmark} \\ \hline
EMA Smoothing & \multicolumn{1}{c}{$\times$} & \multicolumn{1}{c}{$\times$}  & \multicolumn{1}{c}{$\times$} & \multicolumn{1}{c}{$\times$} & \multicolumn{1}{c}{\checkmark} \\ \hline
\end{tabular}
\end{table*}

\section{Preliminaries}

\subsection{Real-world considerations for practical FSSL}
\label{real-world-design}
We describe some of the main considerations when designing practical FSSL systems, such as \textit{privacy leakage} and \textit{client state}. These are general issues for designing standard FL systems, and any semi-supervised approach also has to be cognizant of these to deploy in the real-world. There may exist a trade-off between satisfying these constraints and overall performance, and generalization performance may improve but at the cost of relaxing these considerations. 

\textbf{Privacy Leakage.} We guarantee the same degree of privacy as in standard federated learning~\cite{FedAvg} by ensuring that (1) there is no additional information provided to the server other than client updates (gradients), and (2) there is no communication between clients that involves sharing their updates or local model parameters. 
Existing literature has shown that it is possible to launch attacks given access to model gradients/parameters both in the central learning setting, as in Shokri et al.~\cite{shokri2017membership} and Carlini et al.~\cite{carlini2021membership}; and in the federated learning setting, such as in Geiping et al.~\cite{geiping2020inverting}.
Not all contemporary FSSL approaches are limited to these attacks. In FedMatch~\cite{FedMatch} for instance, the inter-client consistency loss requires the server to store all client models and send $H$ helper agents (the $H$ nearest client models) to each client. This poses an additional privacy vulnerability, as malicious clients could launch attacks when provided access to other client models, with the goal of impairing model training or inferring other clients' data. 

\textbf{Client state.} In large-scale deployments of cross-device FL, most clients do not participate more than once during training, and thus the design of cross-device FSSL algorithms should be stateless. Concretely, stateless clients do not retain any information pertaining to previous rounds~\cite{kairouz2021advances,singhal2021federated}. Not all previous algorithms satisfy this desideratum. For instance, FedCon~\cite{long2021fedcon} assumes that clients retain their projector network across multiple rounds, and in FedSEAL~\cite{bian2021fedseal} clients utilize historical information through self-ensembling across multiple rounds. These approaches assume stateful clients, which may be problematic in a real-world setting.


\begin{figure*}[t]
  \centering
  \includegraphics[width=0.95\linewidth]{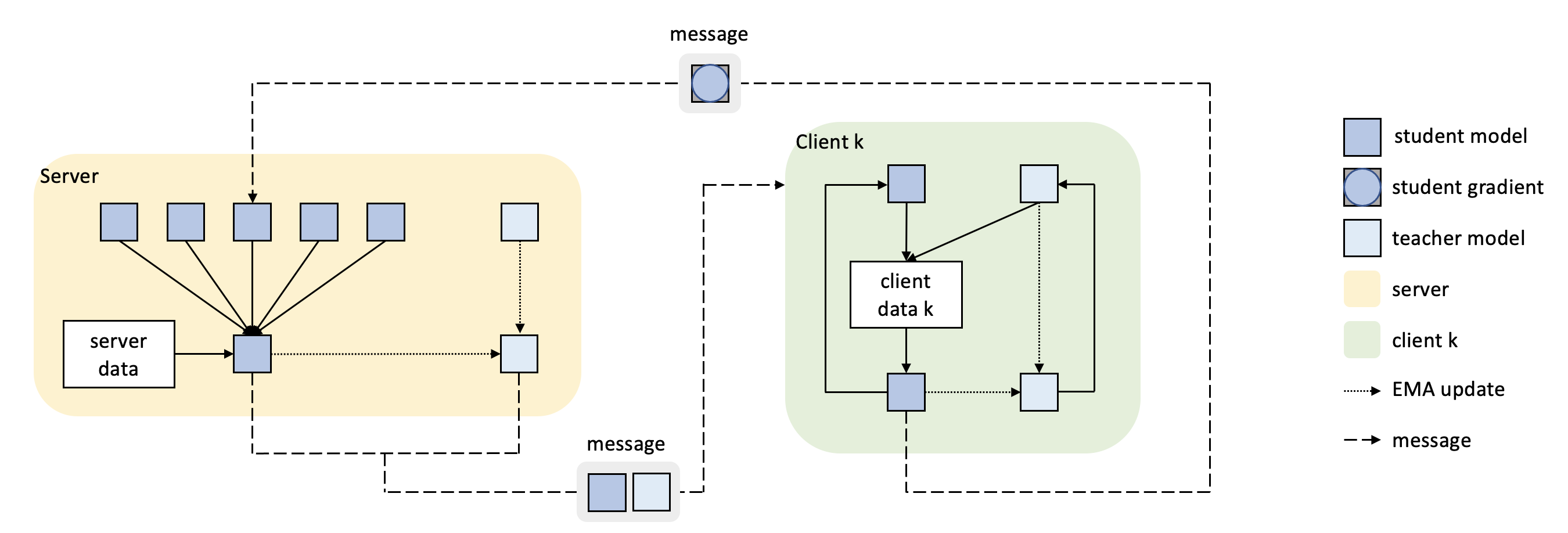}
   \caption{Overview of teacher-student EMA modeling for FSSL including transport of teacher/student models between clients and central server. We can reduce communication cost by sending only the student model from each client to the server.}
   \label{ts-ema-overview}
\end{figure*}

\subsection{FSSL with teacher-student EMA models} \label{fssl-ema-desc}
\textbf{EMA teacher-student model.}
Inspired by Mean Teacher~\cite{tarvainen2017mean} and Unbiased Teacher~\cite{liu2021unbiased}, we leverage the EMA as a way to generate more stable pseudo-labels. 
Unlike the FixMatch method~\cite{sohn2020fixmatch}, where a single model both generates pseudo-labels and trains on them, in the teacher-student framework with EMA, we maintain two separate models, where $\theta_{t}$ and $\theta_{s}$ are the learnable parameters of the teacher and the student model respectively. An EMA update is performed after each update of the (global) student model.
For simplicity, EMA update can be defined as (where $\alpha$ is the EMA ratio.):
\begin{equation}\label{ema_eq}
\theta_{t}  \leftarrow \alpha \theta_{t} + (1-\alpha) \theta_{s}
\end{equation}

\noindent EMA has found applications in the Semi-supervised~\cite{tarvainen2017mean,liu2021unbiased} and Self-supervised learning settings~\cite{he2020momentum,grill2020bootstrap,caron2020unsupervised}. 
In the semi-supervised learning 
context, this is particularly useful, because the teacher model that is updated via EMA has been shown to generate stable and robust pseudo-labels, especially in real-life scenarios where pseudo-labels might be biased and imbalanced~\cite{liu2021unbiased}. While there has been success in using such a method, we, for the first time, demonstrate its effectiveness in the FSSL context.

\textbf{Teacher student EMA models in FL settings}. During EMA updates in a non-federated setting, the teacher model is updated after every training batch and is able to stay synchronized with the student model, while being relatively more robust to fluctuations which could potentially affect its performance. In a federated setting training occurs on each client, and might also occur at the central server in addition to aggregation, particularly in the labels-at-server setting. This implies that we need to maintain a separate teacher model for each client device which we refer to as the local teacher models $\{\teacher^k_{t+1}, k\in L_t\}$, in addition to the global teacher model $\teacher_t$ which resides on the server. 

Similar to the teacher, we maintain local student models $\{\student^k_{t+1}, k\in L_t\}$ as well as a global student model $\student_t$. For each client $k$, the local teacher model $T^{k}_{t}$ gets updated through EMA based on the local student model $S^{k}_{t}$ which is located on the same device. The student models participate in the standard FL protocols, where they are trained using local client data and aggregated to form the server-side global student model. The global teacher $T_t$ is updated from the global student model $S_t$ through EMA as in Equation~\ref{ema_eq} at the end of the $t$-th round. The salient feature of this protocol is that we transmit the global teacher model to all clients at the end of each round to keep the global and local teachers synchronized. For the remainder of the paper, we refer to this approach as \textit{\freqema}.


\subsection{Practical challenges with EMA models in FL}\label{practical-ema}
In Section~\ref{fssl-ema-desc} we have described a potential approach to implementing teacher-student EMA models in a federated learning setting. This adaptation of EMA is the closest to the server-side (central) update mechanism - the teacher and student models are updated after every batch of data on their respective devices, and in every $t$-th communication round the updates from the local student models $\student^k_t$ on each $k$-th client are sent to the server along with the global teacher model being sent to each client for synchronization. This communication is in addition to the global student model being sent to the clients for local training, as well as local students being sent to the server for aggregation. Figure~\ref{ts-ema-overview} shows the teacher and student model transmission between clients and the server. The two-way additional communication overhead has the disadvantage of requiring a higher wall-clock time to reach a target performance in a production setting, due to the additional overhead of transmitting two models from the clients to the server, and in the other direction.

\begin{figure}[t]
\centering
\includegraphics[width=\linewidth]{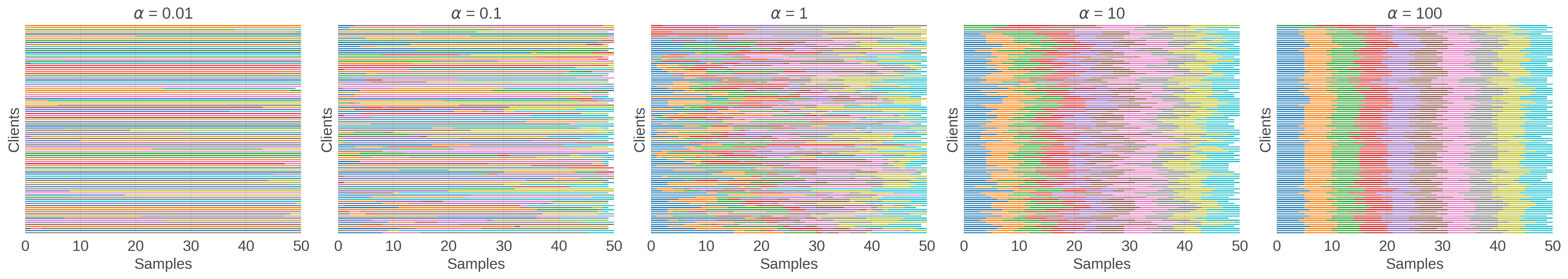}
\caption{
Distribution of each client's labeled examples over CIFAR-10 classes with increasing non-IIDness. Each row represents one client's data and each color represents data from one class. At large $\alpha$, data is mostly IID. As we decrease $\alpha$, clients exhibit label skew, and at small $\alpha$, each client only has data from 1-2 classes.
}
\label{fig:iidness_bar_labeled}
\end{figure}

\setcounter{LaS}{0}
\setcounter{EMA}{1}
\renewcommand{\modelparam}{\student}

\begin{algorithm}[bht]
\ifLaS{
    \caption{\textbf{Labels-at-Server\ifparallel{(parallel).}{(sequential).}}Client devices $1,\dots, K$, participation rate $C$, client learning rate $\eta_c$, server learning rate $\eta_s$, number of local client epochs $E_c$, number of local server epochs $E_s$, FixMatch parameter $\lambda_u$, FedProx parameter $\mu$\ifEMA{, EMA ratio $\alpha$}{}. In practice, \textbf{ClientUpdate} returns the difference $\modelparam-\modelparam^s$ instead of $\modelparam$.}
    \ifEMA{\label{alg:las_EMA}}{\label{alg:las}}
}{
    \caption{
    \textbf{Labels-at-Client.} 
    Client devices $1,\dots,K$, participation rate $C$, learning rate $\eta$, number of local epochs $E$, FixMatch parameter $\lambda_u$, FedProx parameter $\mu$\ifEMA{, EMA ratio~$\alpha$.}{.} In practice, \textbf{ClientUpdate} returns the difference $\modelparam-\modelparam^s$ instead of $\modelparam$.}
    \ifEMA{\label{alg:lac_EMA}}{\label{alg:lac}}
}
\scriptsize
\begin{algorithmic}

\STATE \textbf{Server executes:}
    \STATE initialize global
    \ifEMA{student model $\student_0$ and teacher model $\teacher_0$}{model $\modelparam_0$}
    \FOR{each round $t = 1, 2, \dots$}
        \STATE $m \leftarrow \max(C \cdot K, 1)$
        \STATE $L_t \leftarrow$ (random set of $m$ clients)
        \IF {$|\kl^\teacher - \alphakl| < |\kl^\student - \alphakl|$}
            \STATE $\teacher_t^\prime \leftarrow \emptyset$
        \ELSE
            \STATE $\teacher_t^\prime \leftarrow \teacher_t$
        \ENDIF
        \FOR {each client $k \in L_t$ \textbf{in parallel}}
            \STATE $\modelparam_{t+1}^k$, $(\kl^\teacher)^k$, $(\kl^\student)^k  \leftarrow \textbf{ClientUpdate}$
            \ifEMA{$(k, \student_t, \teacher_t)$}{$(k, \modelparam_t)$}
        \ENDFOR
        \ifparallel{
            \STATE $\modelparam_{t+1} \leftarrow \textbf{ServerUpdate}(\modelparam_t)$ \COMMENT{in parallel}
        }{}
        \STATE $\modelparam_{t+1} \leftarrow$
            \ifparallel{$\frac{n_s}{n}\modelparam_{t+1} +\frac 1 m \sum_{k \in L_t} \modelparam_{t+1}^k$}{$\frac 1 m \sum_{k \in L_t} \modelparam_{t+1}^k$}
            \ifLaS{}{$\frac 1 m \sum_{k \in L_t} \modelparam_{t+1}^k$}
        \ifparallel{}{\STATE $\modelparam_{t+1} \leftarrow \textbf{ServerUpdate}(\modelparam_{t+1})$}
        \ifEMA{\STATE $\teacher_{t+1} \leftarrow \alpha \teacher_t + (1-\alpha) \student_{t+1}$}{}
        \STATE $\kl^\teacher \leftarrow \frac 1 m \sum_{k \in L_t} (\kl^\teacher)^k$
        \STATE $\kl^\student \leftarrow \frac 1 m \sum_{k \in L_t} (\kl^\student)^k$
    \ENDFOR

\ifLaS{
    \STATE
    \STATE \textbf{ServerUpdate$(\modelparam)$:}
        \FOR {each local epoch $e = 1, 2, ... E_s$}
            \FOR {labeled batch $s \in D$}
                \STATE $\modelparam \leftarrow \modelparam - \eta_s \cdot \nabla \suploss(\modelparam; s)$
            \ENDFOR
        \ENDFOR
        \STATE return $\modelparam$
}{}

\STATE
\STATE \textbf{ClientUpdate\ifEMA{$(k, \student, \teacher)$}{$(k, \modelparam)$}:}
    \STATE $\kl^\teacher \leftarrow 0$
    \STATE $\kl^\student \leftarrow 0$
    \FOR {each local epoch $e = 1, 2, \dots,E_c$}
        \FOR {\ifLaS{unlabeled batch $u \in D$}{labeled batch $s \in D_s$ and unlabeled batch $u \in D_u$} } 
        \ifLaS{
            \STATE $\modelparam \leftarrow \modelparam - \eta \cdot \nabla (\lambda_u(\unsuploss$ 
            \ifEMA{$(\student, \teacher; u)$}{$(\modelparam; u)$}
             $) + \frac{\mu}{2} || \modelparam - \modelparam^s||^2 )$
        }{
            \ifEMA{\STATE $\modelparam \leftarrow \modelparam - \eta \cdot \nabla (\suploss(\modelparam; s) + \lambda_u(\unsuploss (\student, \teacher; u)) + \frac{\mu}{2} || \modelparam - \modelparam^s||^2)$}{\STATE $\modelparam \leftarrow \modelparam - \eta \cdot \nabla (\suploss(\modelparam; s) + \lambda_u(\unsuploss (\modelparam; u)) + \frac{\mu}{2} || \modelparam - \modelparam^s||^2)$}
        }
        \STATE $\kl^\teacher \leftarrow \kl^\teacher + \kl^\teacher(u)$
        \STATE $\kl^\student \leftarrow \kl^\student +  \kl^\student(u)$
        \ENDFOR
    \ENDFOR
    \STATE return $\modelparam$, $\kl^\teacher$, $\kl^\student$ to server
\end{algorithmic}
\end{algorithm}

\noindent
\textbf{Communication savings in teacher-student EMA.} We can reduce the communication cost of the EMA approach in a federated setting by not sending the local teacher model $\teacher^k_t$ from each client to the server after every round.
The student model at each client is updated by optimizing labeled and unlabeled losses through FixMatch~\cite{sohn2020fixmatch} and an additional FedProx~\cite{FedProx} loss term. The updated client student models $\{\student^k_{t+1}, k\in L_t\}$ are aggregated at the server to obtain the global student model $\student_{t+1}$. The global teacher model is updated through EMA from the aggregated global student model, resulting in $\teacher_{t+1}$. Clients utilize their copy of the global teacher model~$\teacher_t$ to generate pseudo-labels in each round. The global teacher does not get updated during client training; nor is it sent to the server along with the local student model updates. We refer to this approach in the rest of the paper as \textit{\vanillaema}. Note that setting $\teacher = \student$ and skipping the EMA update for the global teacher model would reduce to FedProx-FixMatch.

\setcounter{LaS}{1}
\setcounter{EMA}{1}
\renewcommand{\modelparam}{\student}

\section{Proposed Method: \proposedregmethod}
\label{proposed_method}
In this section we describe the proposed \proposedregmethod{} approach, which has two main modifications to the \vanillaema approach described in Section~\ref{practical-ema}: \textit{local teacher adaptation} and \textit{adaptive switching between teacher and student}. We also motivate these modifications and conduct ablation studies to evaluate how these modifications impact generalization performance. 

\textbf{Local Teacher Adaptation.} The \vanillaema approach is communication efficient as it does not send two model updates to the server from the $k$-th client unlike \freqema. However the local teacher model $T^k_t$ at each $k$-th client is not updated through EMA as the local student model $S^k_t$ is trained on batches of client data. This results in sub-optimal performance, particularly when the prediction distribution of the local teacher differs from the true underlying label distribution on the client. In \proposedregmethod{}, we do not send the local teacher updates to the server but continue to update it from the local student with every batch through an EMA update. This enables the teacher model to reap the benefits of EMA smoothing while not increasing the client to server transport cost.

\textbf{Adaptive Switching.} Practical federated learning settings are often non-IID where the client devices differ from each other in label distribution. The client non-IIDness property is perhaps the best-studied of all constraints~\cite{zhu2021federated}. Traditional teacher-student methods train the student model on unlabeled data by generating pseudo-labels with the teacher model. The teacher is updated via EMA and can be considered a temporal ensemble of student models, thereby stabilizing and maintaining the quality of the pseudo-labels. However, solely using the teacher to generate pseudo-labels is often sub-optimal in non-IID FL settings.

In addition to reducing communication costs by only sending the model updates from the student back to the server, in each round, \proposedregmethod dynamically chooses between using the local teacher or student model to generate pseudo-labels based on an IIDness prior $\alphakl$, which is a hyper-parameter that represents the amount of expected imbalance in the client's unlabeled data. For each unlabeled batch, we calculate the KL-divergence between the generated pseudo-label predictions and an uniform distribution for both the teacher and student model. At the end of each round, the server aggregates the divergences from all participating clients into a global KL-divergence for the teacher and student model respectively. If the teacher's mean KL-divergence is closer to $\alphakl$, both the teacher and student model are sent to clients in the next round, and clients use the teacher model to generate pseudo-labels. If the student's mean KL-divergence is closer to $\alphakl$, only the student model is sent to clients in the next round, and clients use the student model to generate pseudo-labels similar to FedProx-FixMatch. We are using the KL-divergence as a proxy metric for pseudo-label quality. 


We refer to the KL-divergence between probability distribution $P$ and the discrete uniform distribution $\mathcal{U}$ over $L$ labels as $D_{KL}(P \parallel \mathcal{U})$. When each client's $l$-th unlabeled batch $B_l = \{x^u_0, x^u_1, ..., x^u_N\}$ contains $N$ examples, we calculate the KL-divergence of both the teacher's prediction distribution $\hat{Y}^T_l$  on weakly augmented data and the student's prediction distribution $\hat{Y}^S_l$ on strongly augmented data.
\begin{align}
    \kl^m &= \frac 1 L \sum_l \kl(\hat{Y}^m_l \parallel \mathcal{U})
\end{align}
where $m \in \{\teacher, \student\}$, $\hat{Y}^m_l$ is the distribution of predictions generated via model $m$ on batch $B_l$ of unlabeled data. $L$ denotes the number of unlabeled batches on the client.

Based on the current round's averaged $\kl^\teacher$, $\kl^\student$, and the hyper-parameter $\alphakl$, the server decides whether to send the teacher model to the clients participating in the next round:
\begin{align}
    \teacher^\prime &= 
    \begin{cases}
        \teacher  & \text{if } |\kl^\teacher - \alphakl| < |\kl^\student - \alphakl|\\
        \emptyset  &  \text{otherwise}
    \end{cases}
\end{align}

We can express supervised loss $\suploss$ and unsupervised loss $\unsuploss$ based on $\student$ and $\teacher^\prime$ as follows:
\begin{align}
    \suploss(S; s) &= \sum_i \mathcal{L}(x^s_i, y_i)\\
    \unsuploss(\student, \teacher^\prime; u) &= 
    \begin{cases}
        \sum_i \mathcal{L}(x^u_i, y^{T^\prime}_i)  & \text{if } \teacher^\prime \neq \emptyset \\
        \sum_i \mathcal{L}(x^u_i, y^S_i)  &  \text{otherwise}
    \end{cases}
\end{align}
where $y^T_i$ and $y^S_i$ are the pseudo-labels generated via the teacher and student model on the weakly augmented version of the $i$-th unlabeled example $x^u_i$ respectively. 

\section{Experiments}
\label{sec-experiments}
\label{exp-sec}
\subsection{Datasets} We conduct experiments and report results on the CIFAR-10 and Fashion MNIST datasets. We follow a similar dataset configuration as the FedMatch~\cite{FedMatch} paper under both the labels-at-client and labels-at-server configuration.

\textbf{CIFAR-10.} We follow FedMatch's custom split of the full dataset into 2,000 validation, 2,000 test, and 56,000 training images\footnote{While the FedMatch paper~\cite{FedMatch} states that they split the data into 3,000 validation, 3,000 test, and 54,000 training images, we chose to follow the split in their code at \url{https://github.com/wyjeong/FedMatch/blob/main/config.py\#L53-L54}}. This is a common configuration followed by existing literature for comparison with state-of-the-art approaches. The configurations support 10 target classes and 100 clients with batch-IID and batch-nonIID splits under both labels-at-client and labels-at-server settings, which results in four scenarios. Under labels-at-client, each client has 5 labeled instances per class, while under labels-at-server, the labeled data (5000 instances) resides at the server with completely unlabeled data at the clients. 

\textbf{Fashion MNIST.} Similar to CIFAR-10, we also implement the same data configuration for the Fashion MNIST dataset as provided in the FedMatch~\cite{FedMatch} paper. The authors however have not released their Fashion MNIST code, so we have resorted to a self-implementation of their streaming non-IID setting which follows the paper as closely as possible. The entire dataset of 70000 images is split into 63000 training, 3500 validation and 3500 testing images. We assume that the unlabeled data on each client is imbalanced and streams in at every round when the client is selected, which is the same setting as \textit{Streaming-NonIID} as in the FedMatch implementation. For this dataset, we assume two cases - labels-at-client, where the labeled and unlabeled instances reside at the client, and labels-at-server, where the unlabeled instances reside at the client, with labeled data at the server.

\subsection{Experimental setup} 
\noindent
\textbf{CIFAR-10.} For experiments on the CIFAR-10 dataset, all training runs use 100 clients in total, 5 clients per round, and 1 local client epoch per round. For ease of comparison with baseline approaches, we follow FedMatch settings  in our implementation. Data is distributed evenly across 100 clients for labels-at-client, i.e. we have 5 labels per class per client for a total of 5,000 labeled examples. The remaining training data is unlabeled, and split either evenly (batch-IID) or with class-imbalance (batch-nonIID) across clients. The setup for labels-at-server setting assumes 100 labeled instances per class on the server, and 550 unlabeled training images per client. 

\textbf{Fashion MNIST.} FSSL training on Fashion MNIST considers 10 clients and 1 local epoch per round, with all clients participating per round. The \textit{Streaming-NonIID} setting assumes the unlabeled data on each client is split into the number of streaming steps (which is 10), with each split being used for training every round. In all experiments, we use weakly augmented images with flip-and-shift in contrast to FedMatch's approach, where they utilize original images. We have observed this to lead to improved training stability and generalization performance, particularly under labels-at-server setting. 


\subsection{Hyper-parameters} For our FedProx-FixMatch, EMA baselines, and \proposedregmethod{} implementations,
we tuned the learning rate, pseudo-label threshold, weight of the unlabeled loss, weight decay, optimizer momentum, and EMA decay (where applicable).
For each method, the best performing hyper-parameters are utilized to run five trial training runs, where each trial has a different seed corresponding to the user selector, model initialization and dataset sharding. For each method, we record the average over five trials along with the standard deviation.

\subsection{Performance under increasing non-IIDness}
\label{non-iidness setup}
\noindent
Practical federated learning settings are often non-IID, where the client devices differ from each other in label distribution. The client non-IIDness property is perhaps the best-studied of all constraints~\cite{zhu2021federated}, and we follow Dirichlet sampling to simulate non-IIDness of both labeled and unlabeled data, as done in Hsu et al.~\cite{hsu2019measuring}. The Dirichlet alpha $\alpha$ controls the level of non-IIDness. For extremely non-IID setting $\alpha$ is low, as $\alpha$ increases, samples are sharded more uniformly across clients. See Figure~\ref{fig:iidness_bar_labeled} for a visualization of samples on each clients under various degrees of non-IIDness ($\alpha$). We choose $\alpha \in \{0.01, 0.05, 0.1, 1.0, 10.0, 100.0\}$ to cover the entire range of non-IIDness adequately. We expect that as non-IIDness increases, performance should decrease as reported in literature~\cite{FedAvg}.

We measure the KL-Divergence (with an uniform distribution as a reference similar to that described in Section~\ref{proposed_method}) of the ground truth label distribution and the teacher-generated pseudo-label distribution for each client and compute the average over all participating clients within each round. For our baseline EMA approaches we conduct a further analysis by 
measuring the ratio of the pseudo-label KL divergence to the ground truth label KL divergence at steady state convergence and comparing different approaches based on this metric.



\begin{figure*}[t]
    \centering
    {{\includegraphics[width=0.47\linewidth]{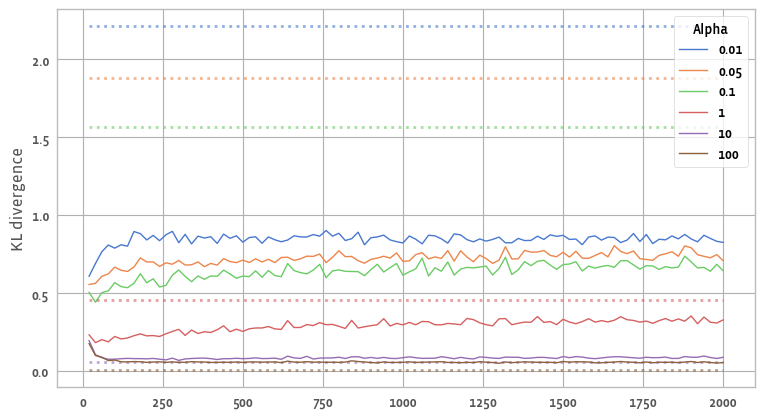}}} 
    \qquad
    {{\includegraphics[width=0.47\linewidth]{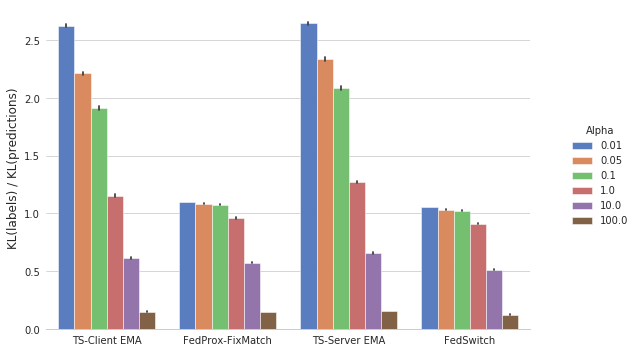}}}
    \caption{Left: KL divergence of the predictions generated by 
    \vanillaema against a uniform distribution throughout training for various degrees of non-IIDness (alpha). Dotted lines are the KL divergence of the labels against a uniform distribution (ground truth). Right: KL divergence of the predictions generated by \proposedregmethod{} divided by the KL divergence of the labels for the last 100 epochs of training (after training has stabilized).
    }
    \label{fig:iidness}
\end{figure*}

\begin{table*}[htbp] \label{cifar10-results}
\caption{Performance of all competing approaches on the CIFAR-10 dataset. For each approach, we report the mean performance, along with standard deviation over five trials and 800 epochs. The best configuration for each approach is obtained through a hyper-parameter search. Results are reported for both Batch-IID and Batch-NonIID settings under labels-at-client and labels-at-server configurations.* refers to existing work that cuts off training at 200 rounds when not yet fully converged.}
\centering
\begin{tabular}{c|cccc}
\hline
                 & \multicolumn{4}{c}{Client/Server Configuration}   \\ \hline
          &\multicolumn{2}{c}{Labels at Client} & \multicolumn{2}{c}{Labels at Server} \\ \hline
 Approach                 & Batch-IID                        & Batch-NonIID                                       & Batch-IID                               & Batch-NonIID                               \\ \hline
FedRGD           &  71.61      &     69.05                                               &        63.32                                 &                   63.24                         \\
FedMatch*         &  52.13$\pm$0.34                                &             52.25$\pm$0.81                                    &           44.95$\pm$0.49                              &                  44.17$\pm$0.19                          \\
FedProx-FixMatch* &   47.20 $\pm$ 0.12                               &             45.55 $\pm$ 0.63                                    &      25.61$\pm$0.32                                  &        9.21$\pm$0.24                                   \\
FedProx-FixMatch (ours) &   90.02$\pm$0.4                               &               89.54$\pm$0.82                                     &      89.8$\pm$0.72                                  &        88.19$\pm$1.10                                \\
\vanillaema          &  89.43$\pm$0.60                                &       85.43$\pm$0.87                                             &      89.15$\pm$0.95 &     73.26$\pm$0.98                           \\
\freqema &  89.94$\pm$0.91                               &     87.41$\pm$0.61                                               &  89.29$\pm$0.63                                &                 78.43$\pm$1.63                     \\
\proposedregmethod  &  \textbf{90.24$\pm$0.60 }                               &     \textbf{89.57$\pm$0.55 }                                             &  \textbf{90.22$\pm$0.49}                                       &                 \textbf{88.91$\pm$0.50}                       \\
\hline
\end{tabular}
\end{table*}

\section{Experimental Results}
\noindent
We have described our experimental setup in Section~\ref{sec-experiments}, which not only includes a comparison of the proposed method \proposedregmethod{} with the state-of-the-art approaches such as FedMatch and FedRGD, but also baseline teacher-student EMA approaches. The experiments also include an analysis of generalization performance under increasing client non-IIDness, and demonstrate that \proposedregmethod{} is relatively more robust compared to baseline approaches when the data distribution on each client is more imbalanced. 
\subsection{Results on CIFAR-10 Dataset}
\label{cifar-10-results}
We benchmark our proposed method \proposedregmethod{} against state-of-the-art methods as well as similar approaches, namely FedProx-FixMatch, FedMatch~\cite{FedMatch}, FedRGD~\cite{FedGRD}, and the EMA baselines (\vanillaema and \freqema).
Within the scope of this paper, we reproduce the FedProx-FixMatch that was originally discussed in FedMatch~\cite{FedMatch}, but instead of cutting off the training at 10 global epochs (200 training rounds) without reaching convergence, we let the method run until convergence (800 epochs, equals 16000 rounds). We ran each experimental setting under both labels-at-client and labels-at-server settings five times with various random dataset seeds and report the average performance and standard deviation. We present results on the CIFAR-10 dataset in Table~\ref{cifar-10-results}. \proposedregmethod{} outperforms the state-of-the-art approaches, FedMatch and FedRGD, along with the baseline EMA variants we have implemented, namely \vanillaema and \freqema.

We also observe that the generalization performance is generally higher for batch-IID splits of the data than for batch-nonIID for almost all methods except FedMatch. Our implementation of FedProx-FixMatch, as well as the EMA variants also does significantly better than the state-of-the-art approaches, FedMatch~\cite{FedMatch} and FedRGD~\cite{FedGRD}. Particularly under labels-at-server, we avoid model collapse/premature convergence by utilizing weak augmentations of the unlabeled data unlike FedMatch. 

\begin{table}[tbp] \label{fashion-mnist-results}
\caption{Performance of all competing approaches on the Fashion-MNIST dataset. For each approach, we report the mean performance, along with standard deviation over five trials and 800 epochs. The best configuration for each approach is obtained through a hyper-parameter search. Results are reported for the Streaming-NonIID settings under both labels-at-client and labels-at-server configurations.}
\centering
\setlength\tabcolsep{1.5pt}
\begin{tabular}{c|cc}
\hline
                 & \multicolumn{2}{c}{Client/Server Configuration} \\ \hline
         & Labels at Client & Labels at Server \\ \hline
      Approach           & Streaming-NonIID       & Streaming-NonIID       \\ \hline
FedMatch         & 77.95$\pm$0.14                     & \textbf{84.15$\pm$0.31}                     \\
FedProx-FixMatch & 62.40$\pm$0.43 & 73.71$\pm$0.32 \\

FedProx-FixMatch(o) & 86.51$\pm$1.05 & 82.77$\pm$1.04                     \\
\vanillaema           &          82.70$\pm$0.67       &        77.49$\pm$5.66\\
\proposedregmethod           & \textbf{88.94$\pm$0.52}                     & 81.72$\pm$2.48\\
\freqema  & 86.48$\pm$0.98                    & 81.15$\pm$3.93                       \\ \hline
\end{tabular}
\end{table}

\subsection{Results on Fashion MNIST Dataset}
\label{fashion-mnist-results-sec}
We compare the performance of \proposedregmethod on the Fashion MNIST dataset with state-of-the-art approaches such as FedMatch, FedRGD as well as other EMA variants. Table~\ref{fashion-mnist-results} shows the average accuracy along with standard deviation over 5 trials for each streaming-nonIID data configuration, such as labels-at-client and labels-at-server. Figure~\ref{convergence-speed}(b) demonstrates the inherent noisiness of this dataset compared to CIFAR-10, which is attributable to not only a non-IID client data setting, but also the temporal nature of the unlabeled data with new samples streaming in on every participating client at each round. The EMA variants are not only able to smoothen the training, but also generalize better at higher epochs since the teacher's pseudo-labels remain relatively robust to temporal variations in the unlabeled data. In Table~\ref{fashion-mnist-results} we present the test performance of \proposedregmethod{} along with the baseline approaches. \proposedregmethod{} obtains an improvement of 11\% accuracy under the Labels-at-Client setting for streaming non-IID data compared to FedMatch, and also outperforms FedProx-FixMatch, both as reported in the FedMatch paper~\cite{FedMatch}, as well as our implementation. Under the Labels-at-Server setting, we also improve on FedProx-FixMatch 
by 2\% test accuracy. Similar to the observations for CIFAR-10, the \freqema approach does better than \vanillaema since we have increased communication as well as increased ability for the local teacher model on the clients to get updated with every batch of data.  

\begin{figure}[t]
  \centering
  \includegraphics[width=\linewidth]{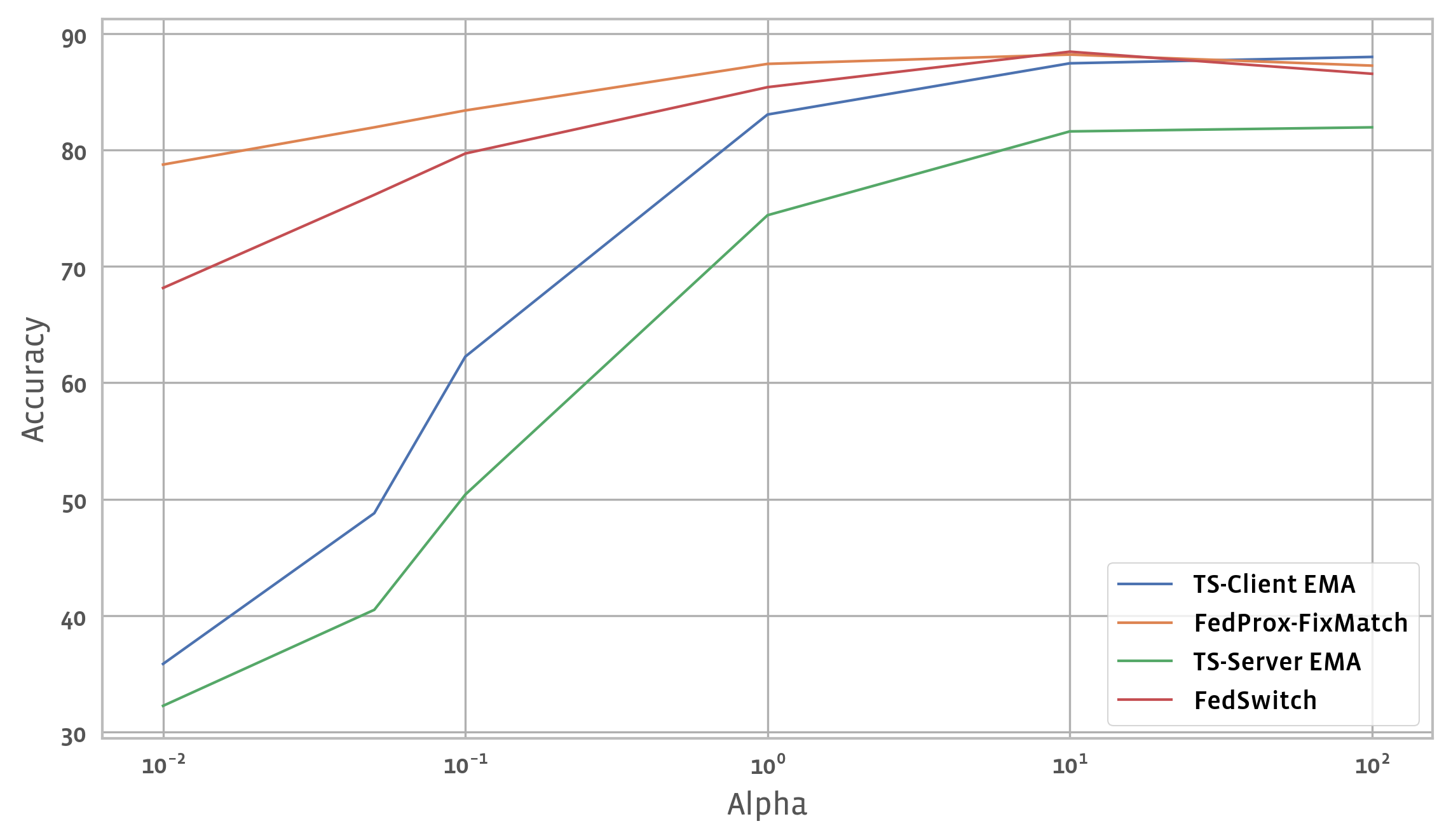} 
   \caption{CIFAR-10 generalization performance of our proposed method \proposedregmethod{} along with baseline approaches at increasing levels of client non-IIDness for Labels at Server (sequential). We note that \proposedregmethod{} is relatively more robust to high levels of client non-IIDness compared to other approaches.}
   \label{cifar10-perf}
\end{figure}

\subsection{Analysis of performance under Non-IIDness}
\label{non-iid-analysis}
We have described in Section~\ref{non-iidness setup} our experimental setup for comparing the proposed method \proposedregmethod along with other teacher-student EMA approaches at different levels of client non-IIDness. Figure~\ref{ts-ema-overview} shows the evaluation accuracy on CIFAR-10 for different approaches (FedProx-Fedmatch, \vanillaema, \freqema and \proposedregmethod) with increasing values of client non-IIDness parameterized by the Dirichlet $\alpha$. We note how \vanillaema and \freqema approaches are susceptible to label imbalance on the clients. This happens as even though each client is imbalanced, the global teacher model is obtained through an aggregation of local teacher models and is biased towards generating pseudo-labels which are relatively more uniform that the true distribution on each client. 

Figure~\ref{fig:iidness} shows how this manifests for different values of Dirichlet $\alpha$. In Figure~\ref{fig:iidness}(a) we plot the variation of the average KL-Divergence between each client's pseudo-label distribution and the uniform distribution with each round as training progresses for the \vanillaema approach. When we have perfect IIDness (balanced labels) on each client, the ground-truth divergence is zero; and on the other extremity ie. each client has samples only from a single class the divergence is $ln(10)=2.302$. 
When the client data is IID, we observe that the teacher can predict pseudo-labels which are uniformly distributed and hence close to the ground truth. However as $\alpha$ decreases, and the data becomes more imbalanced the KL Divergence saturates around 0.75, and the generated pseudo-labels are biased towards an uniform distribution, being unable to match the relatively higher-levels of non-IIDness on the clients. 

 A comparison of how different teacher-student EMA approaches mitigate this issue is presented in Figure~\ref{fig:iidness}(b). For FedProx-FixMatch the ratio of ground truth KL-divergences to the pseudo-label KL-Divergences is close to 1.0 which indicates that we have good quality of predictions. The \vanillaema approach performs the worst, with no adaptation of the local student or teacher model to the client data. When we increase the communication cost (\freqema), we perform relatively better than \vanillaema due to improved ability of the local teacher model on each client to track variation of the student models. \proposedregmethod{} performs similar to FedProx-FixMatch and is relatively unaffected by non-IIDness. The reasons for this are two-fold - firstly, the local student is adapted to the client data which also enables us to update the local teacher model; and secondly, depending on the switching criterion the model being used for pseudo-label generation can adaptively switch between the teacher and the student on each client. This enables the pseudo-labels to match the true client distribution, even under increasing levels of non-IIDness. 

\begin{figure}[t]
    \centering
    {{\includegraphics[width=\linewidth]{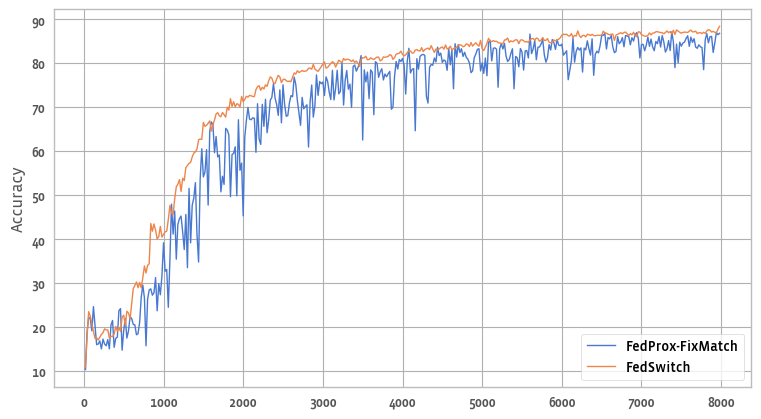}}} \\
    {{\includegraphics[width=\linewidth]{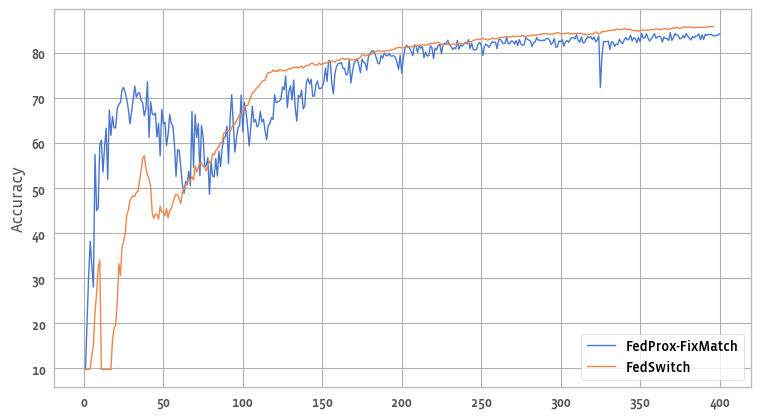}}}
    \caption{Variation of evaluation accuracy with number of global rounds for FedSwitch and FedProx-FixMatch approaches on the CIFAR-10 dataset (batched nonIID) and on Fashion MNIST dataset (streaming nonIID) under labels-at-client settings. \proposedregmethod\ achieves better performance compared to the baseline approaches at fewer rounds, while being more stable during training. 
    }
    \label{convergence-speed}
\end{figure}

\section{Conclusion}
\noindent
In this paper we have proposed \proposedregmethod, a novel teacher-student EMA framework for semi-supervised learning in a federated setting. We show that the problem of integrating EMA updates when data resides on different client devices is non-trivial, and requires synchronization between global student/teacher models, and local ones residing on each client. When evaluated on two datasets, CIFAR-10 and Fashion MNIST for the federated image classification task, \proposedregmethod{} obtains performance superior to existing state-of-the-art approaches while utilizing teacher student EMA updates with a tractable communication cost. Unlike existing approaches such as FedMatch, our proposed method also satisfies the necessary requirements such as privacy and client statelessness for deployment to practical federated settings. In future work, we plan to extend \proposedregmethod{} to incorporate techniques such as Test-Time Adaptation with applications to other federated computer vision tasks, such as image segmentation and instance-level object detection.

\bibliography{main}

\begin{thebibliography}{40}
\providecommand{\natexlab}[1]{#1}
\providecommand{\url}[1]{\texttt{#1}}
\expandafter\ifx\csname urlstyle\endcsname\relax
  \providecommand{\doi}[1]{doi: #1}\else
  \providecommand{\doi}{doi: \begingroup \urlstyle{rm}\Url}\fi

\bibitem[Berthelot et~al.(2019)Berthelot, Carlini, Goodfellow, Papernot,
  Oliver, and Raffel]{berthelot2019mixmatch}
Berthelot, D., Carlini, N., Goodfellow, I., Papernot, N., Oliver, A., and
  Raffel, C.~A.
\newblock Mixmatch: A holistic approach to semi-supervised learning.
\newblock In \emph{NeurIPS}, pp.\  5049--5059, 2019.

\bibitem[Bian et~al.(2021)Bian, Fu, and Xu]{bian2021fedseal}
Bian, J., Fu, Z., and Xu, J.
\newblock Fedseal: Semi-supervised federated learning with self-ensemble
  learning and negative learning.
\newblock \emph{arXiv preprint arXiv:2110.07829}, 2021.

\bibitem[Carlini et~al.(2021)Carlini, Chien, Nasr, Song, Terzis, and
  Tramer]{carlini2021membership}
Carlini, N., Chien, S., Nasr, M., Song, S., Terzis, A., and Tramer, F.
\newblock Membership inference attacks from first principles.
\newblock \emph{arXiv preprint arXiv:2112.03570}, 2021.

\bibitem[Caron et~al.(2020)Caron, Misra, Mairal, Goyal, Bojanowski, and
  Joulin]{caron2020unsupervised}
Caron, M., Misra, I., Mairal, J., Goyal, P., Bojanowski, P., and Joulin, A.
\newblock Unsupervised learning of visual features by contrasting cluster
  assignments.
\newblock \emph{Advances in Neural Information Processing Systems},
  33:\penalty0 9912--9924, 2020.

\bibitem[Dwork et~al.(2014)Dwork, Roth, et~al.]{dwork2014algorithmic}
Dwork, C., Roth, A., et~al.
\newblock The algorithmic foundations of differential privacy.
\newblock \emph{Found. Trends Theor. Comput. Sci.}, 9\penalty0 (3-4):\penalty0
  211--407, 2014.

\bibitem[Geiping et~al.(2020)Geiping, Bauermeister, Dr{\"o}ge, and
  Moeller]{geiping2020inverting}
Geiping, J., Bauermeister, H., Dr{\"o}ge, H., and Moeller, M.
\newblock Inverting gradients---how easy is it to break privacy in federated
  learning?
\newblock \emph{Advances in Neural Information Processing Systems},
  33:\penalty0 16937--16947, 2020.

\bibitem[Grill et~al.(2020)Grill, Strub, Altch{\'e}, Tallec, Richemond,
  Buchatskaya, Doersch, Pires, Guo, Azar, et~al.]{grill2020bootstrap}
Grill, J.-B., Strub, F., Altch{\'e}, F., Tallec, C., Richemond, P.~H.,
  Buchatskaya, E., Doersch, C., Pires, B.~A., Guo, Z.~D., Azar, M.~G., et~al.
\newblock Bootstrap your own latent: A new approach to self-supervised
  learning.
\newblock \emph{arXiv preprint arXiv:2006.07733}, 2020.

\bibitem[Guillaumin et~al.(2010)Guillaumin, Verbeek, and
  Schmid]{guillaumin2010multimodal}
Guillaumin, M., Verbeek, J., and Schmid, C.
\newblock Multimodal semi-supervised learning for image classification.
\newblock In \emph{2010 IEEE Computer society conference on computer vision and
  pattern recognition}, pp.\  902--909. IEEE, 2010.

\bibitem[Guo et~al.(2019)Guo, Mao, and Zhang]{guo2019mixup}
Guo, H., Mao, Y., and Zhang, R.
\newblock Mixup as locally linear out-of-manifold regularization.
\newblock In \emph{Proceedings of the AAAI Conference on Artificial
  Intelligence (AAAI)}, volume~33, pp.\  3714--3722, 2019.

\bibitem[He et~al.(2020)He, Fan, Wu, Xie, and Girshick]{he2020momentum}
He, K., Fan, H., Wu, Y., Xie, S., and Girshick, R.
\newblock Momentum contrast for unsupervised visual representation learning.
\newblock In \emph{Proceedings of the IEEE/CVF Conference on Computer Vision
  and Pattern Recognition (CVPR)}, pp.\  9729--9738, 2020.

\bibitem[Hendrycks et~al.(2020)Hendrycks, Mu, Cubuk, Zoph, Gilmer, and
  Lakshminarayanan]{hendrycks2020augmix}
Hendrycks, D., Mu, N., Cubuk, E.~D., Zoph, B., Gilmer, J., and
  Lakshminarayanan, B.
\newblock {AugMix}: A simple data processing method to improve robustness and
  uncertainty.
\newblock \emph{Proceedings of the International Conference on Learning
  Representations (ICLR)}, 2020.

\bibitem[Hsu et~al.(2019)Hsu, Qi, and Brown]{hsu2019measuring}
Hsu, T.-M.~H., Qi, H., and Brown, M.
\newblock Measuring the effects of non-identical data distribution for
  federated visual classification.
\newblock \emph{arXiv preprint arXiv:1909.06335}, 2019.

\bibitem[Jeong et~al.(2021)Jeong, Yoon, Yang, and Hwang]{FedMatch}
Jeong, W., Yoon, J., Yang, E., and Hwang, S.~J.
\newblock Federated semi-supervised learning with inter-client consistency {\&}
  disjoint learning.
\newblock In \emph{International Conference on Learning Representations}, 2021.
\newblock URL \url{https://openreview.net/forum?id=ce6CFXBh30h}.

\bibitem[Kairouz et~al.(2021)Kairouz, McMahan, Avent, Bellet, Bennis, Bhagoji,
  Bonawitz, Charles, Cormode, Cummings, et~al.]{kairouz2021advances}
Kairouz, P., McMahan, H.~B., Avent, B., Bellet, A., Bennis, M., Bhagoji, A.~N.,
  Bonawitz, K., Charles, Z., Cormode, G., Cummings, R., et~al.
\newblock Advances and open problems in federated learning.
\newblock \emph{Foundations and Trends{\textregistered} in Machine Learning},
  14\penalty0 (1--2):\penalty0 1--210, 2021.

\bibitem[Karimireddy et~al.(2020)Karimireddy, Kale, Mohri, Reddi, Stich, and
  Suresh]{karimireddy2020scaffold}
Karimireddy, S.~P., Kale, S., Mohri, M., Reddi, S., Stich, S., and Suresh,
  A.~T.
\newblock Scaffold: Stochastic controlled averaging for federated learning.
\newblock In \emph{International Conference on Machine Learning}, pp.\
  5132--5143. PMLR, 2020.

\bibitem[Kone{\v{c}}n{\`y} et~al.(2016)Kone{\v{c}}n{\`y}, McMahan, Yu,
  Richt{\'a}rik, Suresh, and Bacon]{konevcny2016federated}
Kone{\v{c}}n{\`y}, J., McMahan, H.~B., Yu, F.~X., Richt{\'a}rik, P., Suresh,
  A.~T., and Bacon, D.
\newblock Federated learning: Strategies for improving communication
  efficiency.
\newblock \emph{arXiv preprint arXiv:1610.05492}, 2016.

\bibitem[Kuo et~al.(2020)Kuo, Ma, Huang, and Kira]{kuo2020featmatch}
Kuo, C.-W., Ma, C.-Y., Huang, J.-B., and Kira, Z.
\newblock Featmatch: Feature-based augmentation for semi-supervised learning.
\newblock In \emph{Proceedings of the European Conference on Computer Vision
  (ECCV)}, 2020.

\bibitem[Kuznetsova et~al.(2020)Kuznetsova, Rom, Alldrin, Uijlings, Krasin,
  Pont-Tuset, Kamali, Popov, Malloci, Kolesnikov, et~al.]{kuznetsova2020open}
Kuznetsova, A., Rom, H., Alldrin, N., Uijlings, J., Krasin, I., Pont-Tuset, J.,
  Kamali, S., Popov, S., Malloci, M., Kolesnikov, A., et~al.
\newblock The open images dataset v4.
\newblock \emph{International Journal of Computer Vision}, 128\penalty0
  (7):\penalty0 1956--1981, 2020.

\bibitem[Laine \& Aila(2017)Laine and Aila]{laine2016temporal}
Laine, S. and Aila, T.
\newblock Temporal ensembling for semi-supervised learning.
\newblock In \emph{ICLR}, 2017.

\bibitem[Li et~al.(2020)Li, Sahu, Zaheer, Sanjabi, Talwalkar, and
  Smith]{FedProx}
Li, T., Sahu, A.~K., Zaheer, M., Sanjabi, M., Talwalkar, A., and Smith, V.
\newblock Federated optimization in heterogeneous networks.
\newblock \emph{Proceedings of Machine Learning and Systems}, 2:\penalty0
  429--450, 2020.

\bibitem[Lin et~al.(2014)Lin, Maire, Belongie, Hays, Perona, Ramanan,
  Doll{\'a}r, and Zitnick]{lin2014microsoft}
Lin, T.-Y., Maire, M., Belongie, S., Hays, J., Perona, P., Ramanan, D.,
  Doll{\'a}r, P., and Zitnick, C.~L.
\newblock Microsoft coco: Common objects in context.
\newblock In \emph{ECCV}, 2014.

\bibitem[Liu et~al.(2021)Liu, Ma, He, Kuo, Chen, Zhang, Wu, Kira, and
  Vajda]{liu2021unbiased}
Liu, Y.-C., Ma, C.-Y., He, Z., Kuo, C.-W., Chen, K., Zhang, P., Wu, B., Kira,
  Z., and Vajda, P.
\newblock Unbiased teacher for semi-supervised object detection.
\newblock In \emph{Proceedings of the International Conference on Learning
  Representations (ICLR)}, 2021.

\bibitem[Long et~al.(2020)Long, Che, Wang, Ye, Luo, Wu, Xiao, and
  Ma]{DBLP:journals/corr/abs-2012-03292}
Long, Z., Che, L., Wang, Y., Ye, M., Luo, J., Wu, J., Xiao, H., and Ma, F.
\newblock Fedsemi: An adaptive federated semi-supervised learning framework.
\newblock \emph{CoRR}, abs/2012.03292, 2020.
\newblock URL \url{https://arxiv.org/abs/2012.03292}.

\bibitem[Long et~al.(2021)Long, Wang, Wang, Xiao, and Ma]{long2021fedcon}
Long, Z., Wang, J., Wang, Y., Xiao, H., and Ma, F.
\newblock Fedcon: A contrastive framework for federated semi-supervised
  learning.
\newblock \emph{arXiv preprint arXiv:2109.04533}, 2021.

\bibitem[McMahan et~al.(2017)McMahan, Moore, Ramage, Hampson, and Aguera~y
  Arcas]{FedAvg}
McMahan, H.~B., Moore, E., Ramage, D., Hampson, S., and Aguera~y Arcas, B.
\newblock Communication-efficient learning of deep networks from decentralized
  data.
\newblock In \emph{Proceedings of the 20th International Conference on
  Artificial Intelligence and Statistics (AISTATS)}, 2017.

\bibitem[Miyato et~al.(2018)Miyato, Maeda, Koyama, and
  Ishii]{miyato2018virtual}
Miyato, T., Maeda, S.-i., Koyama, M., and Ishii, S.
\newblock Virtual adversarial training: a regularization method for supervised
  and semi-supervised learning.
\newblock \emph{IEEE transactions on pattern analysis and machine intelligence
  (PAMI)}, 41\penalty0 (8):\penalty0 1979--1993, 2018.

\bibitem[Sajjadi et~al.(2016)Sajjadi, Javanmardi, and
  Tasdizen]{sajjadi2016regularization}
Sajjadi, M., Javanmardi, M., and Tasdizen, T.
\newblock Regularization with stochastic transformations and perturbations for
  deep semi-supervised learning.
\newblock In \emph{Advances in Neural Information Processing Systems
  (NeurIPS)}, pp.\  1163--1171, 2016.

\bibitem[Shokri et~al.(2017)Shokri, Stronati, Song, and
  Shmatikov]{shokri2017membership}
Shokri, R., Stronati, M., Song, C., and Shmatikov, V.
\newblock Membership inference attacks against machine learning models.
\newblock In \emph{2017 IEEE symposium on security and privacy (SP)}, pp.\
  3--18. IEEE, 2017.

\bibitem[Singhal et~al.(2021)Singhal, Sidahmed, Garrett, Wu, Rush, and
  Prakash]{singhal2021federated}
Singhal, K., Sidahmed, H., Garrett, Z., Wu, S., Rush, J., and Prakash, S.
\newblock Federated reconstruction: Partially local federated learning.
\newblock \emph{Advances in Neural Information Processing Systems}, 34, 2021.

\bibitem[Sohn et~al.(2020)Sohn, Berthelot, Li, Zhang, Carlini, Cubuk, Kurakin,
  Zhang, and Raffel]{sohn2020fixmatch}
Sohn, K., Berthelot, D., Li, C.-L., Zhang, Z., Carlini, N., Cubuk, E.~D.,
  Kurakin, A., Zhang, H., and Raffel, C.
\newblock Fixmatch: Simplifying semi-supervised learning with consistency and
  confidence.
\newblock In \emph{NeurIPS}, 2020.

\bibitem[Sorokin \& Forsyth(2008)Sorokin and Forsyth]{mturk}
Sorokin, A. and Forsyth, D.
\newblock Utility data annotation with amazon mechanical turk.
\newblock In \emph{2008 IEEE Computer Society Conference on Computer Vision and
  Pattern Recognition Workshops}, pp.\  1--8, 2008.
\newblock \doi{10.1109/CVPRW.2008.4562953}.

\bibitem[Tarvainen \& Valpola(2017)Tarvainen and Valpola]{tarvainen2017mean}
Tarvainen, A. and Valpola, H.
\newblock Mean teachers are better role models: Weight-averaged consistency
  targets improve semi-supervised deep learning results.
\newblock In \emph{Advances in neural information processing systems
  (NeurIPS)}, pp.\  1195--1204, 2017.

\bibitem[Wei et~al.(2020)Wei, Li, Ding, Ma, Yang, Farokhi, Jin, Quek, and
  Poor]{wei2020federated}
Wei, K., Li, J., Ding, M., Ma, C., Yang, H.~H., Farokhi, F., Jin, S., Quek,
  T.~Q., and Poor, H.~V.
\newblock Federated learning with differential privacy: Algorithms and
  performance analysis.
\newblock \emph{IEEE Transactions on Information Forensics and Security},
  15:\penalty0 3454--3469, 2020.

\bibitem[Xie et~al.(2020)Xie, Luong, Hovy, and Le]{xie2020self}
Xie, Q., Luong, M.-T., Hovy, E., and Le, Q.~V.
\newblock Self-training with noisy student improves imagenet classification.
\newblock In \emph{CVPR}, 2020.

\bibitem[Xu et~al.(2021)Xu, Zhang, Hu, Wang, Wang, Wei, Bai, and
  Liu]{xu2021end}
Xu, M., Zhang, Z., Hu, H., Wang, J., Wang, L., Wei, F., Bai, X., and Liu, Z.
\newblock End-to-end semi-supervised object detection with soft teacher.
\newblock \emph{arXiv preprint arXiv:2106.09018}, 2021.

\bibitem[Yu et~al.(2019)Yu, Wu, Ma, and Zhu]{yu2019tangent}
Yu, B., Wu, J., Ma, J., and Zhu, Z.
\newblock Tangent-normal adversarial regularization for semi-supervised
  learning.
\newblock In \emph{Proceedings of the IEEE Conference on Computer Vision and
  Pattern Recognition (CVPR)}, pp.\  10676--10684, 2019.

\bibitem[Yun et~al.(2019)Yun, Han, Oh, Chun, Choe, and Yoo]{yun2019cutmix}
Yun, S., Han, D., Oh, S.~J., Chun, S., Choe, J., and Yoo, Y.
\newblock Cutmix: Regularization strategy to train strong classifiers with
  localizable features.
\newblock In \emph{Proceedings of the IEEE International Conference on Computer
  Vision (ICCV)}, pp.\  6023--6032, 2019.

\bibitem[Zhang et~al.(2018)Zhang, Cisse, Dauphin, and
  Lopez-Paz]{zhang2018mixup}
Zhang, H., Cisse, M., Dauphin, Y.~N., and Lopez-Paz, D.
\newblock mixup: Beyond empirical risk minimization.
\newblock In \emph{Proc. International Conference on Learning Representations
  (ICLR)}, 2018.

\bibitem[Zhang et~al.(2021)Zhang, Yang, Yao, Yan, Gonzalez, Ramchandran, and
  Mahoney]{FedGRD}
Zhang, Z., Yang, Y., Yao, Z., Yan, Y., Gonzalez, J.~E., Ramchandran, K., and
  Mahoney, M.~W.
\newblock Improving semi-supervised federated learning by reducing the gradient
  diversity of models.
\newblock \emph{IEEE International Conference on Big Data (Big Data)}, 2021.

\bibitem[Zhu et~al.(2021)Zhu, Xu, Liu, and Jin]{zhu2021federated}
Zhu, H., Xu, J., Liu, S., and Jin, Y.
\newblock Federated learning on non-iid data: A survey.
\newblock \emph{Neurocomputing}, 465:\penalty0 371--390, 2021.

\end{thebibliography}
\bibliographystyle{icml2022}





\end{document}
